\pdfoutput=1  
\documentclass{article}

\usepackage{arxiv}

\usepackage[numbers]{natbib}

\usepackage[utf8]{inputenc} 
\usepackage[T1]{fontenc}    
\usepackage{amsmath}
\usepackage{amsfonts}       
\usepackage{physics}
\usepackage{nicefrac}       
\usepackage{microtype}      
\usepackage{booktabs}       
\usepackage{multirow}
\usepackage{graphicx}
\usepackage{subcaption}
\usepackage{xcolor}         
\usepackage{url}            
\usepackage{hyperref}       
\hypersetup{
    colorlinks=true,
    linkcolor=blue,
    urlcolor=blue,
    citecolor=blue
}

\newcommand{\calK}{\mathcal{K}}
\newcommand{\calM}{\mathcal{M}}
\newcommand{\calR}{\mathcal{R}}

\newcommand{\MLLM}{\operatorname{MLLM}}
\newcommand{\Refl}{\operatorname{Refl}}
\newcommand{\Synth}{\operatorname{Synth}}
\newcommand{\Cat}{\operatorname{Cat}}
\newcommand{\Emb}{\operatorname{Emb}}
\newcommand{\Verify}{\operatorname{Verify}}
\newcommand{\concepts}{\operatorname{concepts}}

\newcommand{\green}[1]{\textcolor{green}{#1}}
\newcommand{\red}[1]{\textcolor{red}{#1}}

\newcommand{\cmark}{\textcolor{green!55!black}{\ensuremath{\checkmark}}}
\newcommand{\xmark}{\textcolor{red!70!black}{\ensuremath{\times}}}

\usepackage[most]{tcolorbox}
\tcbset{
    promptboxbase/.style={
        enhanced,
        breakable,
        sharp corners=south,
        boxrule=0.6pt,
        left=8pt, right=8pt, top=6pt, bottom=6pt,
        fonttitle=\bfseries\sffamily\small,
        coltitle=white,
        attach boxed title to top left={yshift=-2mm, xshift=4mm},
        boxed title style={
            sharp corners,
            boxrule=0pt,
            top=2pt, bottom=2pt, left=6pt, right=6pt,
        },
    },
}
\newtcolorbox{promptbox}[1]{
    promptboxbase,
    colback=gray!5,
    colframe=black!70,
    colbacktitle=black!70,
    title={#1},
}
\newtcolorbox{promptboxsys}[1]{
    promptboxbase,
    colback=blue!3,
    colframe=blue!55!black,
    colbacktitle=blue!55!black,
    title={#1},
}
\newtcolorbox{promptboxuser}[1]{
    promptboxbase,
    colback=orange!3,
    colframe=orange!70!black,
    colbacktitle=orange!70!black,
    title={#1},
}

\newcommand{\ph}[1]{\textcolor{purple!70!black}{\texttt{\{\{#1\}\}}}}

\title{Dual-Grained Agent Memory and Shapley Context Attribution for Multimodal Agentic Learner}

\author{%
  Jieke Wang \\ 
  UC Merced \\
  \texttt{jwang450@ucmerced.edu} \\
  \And
  Tiancheng Shen \\
  UC Merced \\
  \texttt{tianchengshen@ucmerced.edu} \\
  \AND
  Yibo Yang \\
  Shanghai Jiao Tong University \\
  \texttt{yibo.yang93@gmail.com} \\
  \And
  Ming-Hsuan Yang \\
  UC Merced \\
  \texttt{myang37@ucmerced.edu}
}

\begin{document}

\begingroup
\renewcommand{\today}{}
\maketitle
\endgroup

\begin{abstract}
Frontier multimodal large language models (MLLMs) deliver impressive perception yet still falter on scientific and mathematical reasoning. Parameter-level adaptation is unavailable for closed-weight or on-device backbones, and stateless prompting forfeits any compounding benefit from problems already solved. We propose \textbf{DG-Mem}, a dual-grained agentic memory framework that augments a frozen MLLM with a non-parametric, externally stored memory built once from training-time rollouts and consulted read-only at test time.
Motivated by the Complementary Learning Systems (CLS) account of human memory, DG-Mem factors its store into an instance-grounded exemplar memory and a category-level schema memory of IF-THEN rules, with a transient reflection store mediating their construction so that schemas are synthesized only from abstract reflections, never from exemplar text.
Two design choices distinguish DG-Mem: an online concept categorizer that grows the category space incrementally during training rather than committing to a predefined taxonomy, and a Shapley context attribution procedure that decomposes correctness across the entire retrieved rule set and yields a per-rule utility that re-weights retrieval at test time.
The pipeline introduces no gradient updates and is deployable on closed-weight or on-device backbones. Across MathVista, MMMU, and MMMU-Pro on four open-weight and proprietary backbones (Qwen3.5-27B, Qwen3.5-122B-A10B, GPT-5-Nano, Gemini-3-Flash), DG-Mem improves consistently over no-memory and competitive memory baselines.
\end{abstract}

\section{Introduction}
\label{sec:intro}

Frontier multimodal large language models (MLLMs) now answer a wide range of vision-language questions~\cite{gpt-5, gemini-3-flash, qwen3}, yet their failures on scientific and mathematical reasoning benchmarks remain~\cite{mmmu, mathvista, mmmu-pro}.
Closing this gap via parameter-level adaptation, such as instruction tuning or RLHF, requires gradient access and full supervision. This approach is brittle when the backbone model changes. Furthermore, it is often impossible for proprietary or on-device models.
Prompt-level adaptation, in contrast, is cheap and modular but stateless: the model receives no compounding benefit from problems it has already seen.
Agentic memory frameworks bridge this divide by maintaining an external, non-parametric store that accumulates across problems and conditions a frozen backbone at test time.

Recent agentic memory systems pursue exactly this regime, but most settle for one of two extremes.
\emph{Trajectory and case banks}~\cite{memento, reflexion, jarvis-1, mem0, memorybank} preserve raw episodes and route by similarity or learned values; the resulting context is rich but instance-bound, and aggregation over many cases tends to amplify rather than abstract their idiosyncrasies.
\emph{Single-grain prescriptive playbooks}~\cite{awm, expel, reasoningbank, ace, vilomem} swing to the other extreme, distilling experience into a flat list of rules; the resulting context is compact but loses the concrete grounding that an unfamiliar new problem often needs.
Systems that do combine grains either pair task- and step-level skills~\cite{d2skill} or pair two error-modality streams of the same abstraction level~\cite{vilomem}, rather than separating instance grounding from cross-case abstraction.
Two design questions are left almost entirely open by these systems: 
(i) how memories should be categorized when the underlying concept space is not known a priori and a fixed taxonomy is either too coarse (e.g. a single ``geometry'' bucket) or too fine (e.g. a unique bucket per concept signature)? and 
(ii) how the credit for a correct answer should be distributed across multiple co-retrieved memories so that retrieval ranking can sharpen over time? Existing per-memory utility signals~\cite{memrl, vilomem} attribute success or failure to a single retrieved item, conflating its contribution with the rest of the context.

A complementary perspective comes from cognitive science.
The Complementary Learning Systems (CLS) account of human memory~\citep{mcclelland1995complementary, kumaran2016learning} holds that flexible behavior arises from two interacting stores: a fast, instance-specific system that supports exemplar-based generalization~\citep{nosofsky1986attention}, and a slow, structured system organized around abstract schemas.
Critically, the two are not built in the same way: episodic traces are progressively integrated offline into reusable schemas, a consolidation step that schemas themselves are known to accelerate~\citep{tse2007schemas}.
Equally relevant is what gets re-activated.
Offline replay is preferentially biased toward rewarded experience~\citep{singer2009rewarded,shohamy2010dopamine}, and the rational analysis of memory holds that retrieval accessibility composes cue-driven match strength with a base-level term that tracks how useful a memory has been historically~\citep{anderson1991reflections}.
Inspired by these claims, we design our agentic memory framework with the granularity separation, schema-mediated consolidation, and value-weighted retrieval.

We propose \textbf{DG-Mem}, a dual-grained agentic memory framework.
A frozen MLLM backbone is paired with a non-parametric store factored into an instance-grounded \emph{exemplar memory} and a category-level \emph{schema memory} of IF-THEN rules; a transient \emph{reflection store} acts as the episodic-trace intermediate from which schemas are synthesized, with exemplar text deliberately excluded from schema synthesis to enforce the consolidation constraint.
To resolve the taxonomy question, an online concept categorizer grows the category space incrementally during training rather than committing to a predefined set, allowing to track the empirical concept distribution of the corpus.
To resolve the credit-assignment question, a second offline pass treats the rules retrieved for each problem as players in a cooperative game and uses \emph{Shapley context attribution} over rule-subset rollouts to compute a per-rule utility; at test time, retrieval composes cue similarity with this utility, instantiating the rational-analysis prescription of similarity weighted by track record.
The entire pipeline runs on a frozen backbone with no gradient updates, making it deployable on closed-weight or on-device models.

Our contributions are summarized as follows:

\begin{itemize}
    \item We propose DG-Mem, an agentic memory framework that enables multimodal agents to learn and evolve. An exemplar memory and a schema memory of IF-THEN rules, with schemas synthesized \emph{exclusively} from a transient reflection store so that exemplar text never contaminates the schema, operationalizing CLS-style consolidation from abstract traces.
    \item We propose Shapley context attribution to handle the credit assignment problem of learned schemas. Rules co-retrieved for a problem are scored as players in a cooperative game. The resulting per-rule utility instantiates the rational-analysis view of accessibility as similarity weighted by historical usefulness.
    \item Across four backbones (Qwen3.5-27B / 122B-A10B~\cite{qwen35}, GPT-5-Nano~\cite{gpt-5}, Gemini-3-Flash~\cite{gemini-3-flash}) and three multimodal benchmarks (MathVista~\cite{mathvista}, MMMU~\cite{mmmu}, MMMU-Pro~\cite{mmmu-pro}), DG-Mem consistently beats No-memory baseline, and both offline and online ViLoMem.
\end{itemize}

\section{Related Works}
\label{sec:related}

Recent works have treated the prompt context as a first-class object to be engineered, accumulated, and evolved~\cite{survey:gao2026a, survey:huang2026rethinkingmemorymechanismsfoundation, survey:hu2025memory}.
Existing methods are usually along two axes: what is stored (granularity) and how it is constructed (regime). 

\paragraph{Granularity.}
\emph{Trajectory and case stores} preserve episodic experience and route through similarities or learned values. 
Memento~\cite{memento} formulates case selection as a soft-Q policy over an episodic case bank. MemRL~\cite{memrl} attaches a TD-learned utility to each retrieved trace. Buffers-based methods~\cite{reflexion, jarvis-1, agentkb, bufferofthoughts} and classical text-memory systems~\cite{mem0, memorybank} keep verbatim reasoning episodes or maintain unstructured notes.
\emph{Single-grain prescriptive playbooks} distill experience into a flat list of generic strategies~\cite{awm, expel, reasoningbank, skillweaver, ace}. ReasoningBank~\cite{reasoningbank} and SkillWeaver~\cite{skillweaver} extract heterogeneous episodes into a single bullet list. ACE~\cite{ace} maintains an itemized playbook updated by a Generator-Reflector-Curator with grow-and-refine de-duplication.
\emph{Dual-grain or structured stores} factor memory into multiple types: ViLoMem~\cite{vilomem} runs visual distractions and logical hallucinations but both at schema-level guideline granularity; D2Skill~\cite{d2skill} stores task-level and step-level skill banks; structured stores such as A-Mem~\cite{amem}, Zep~\cite{zep}, and G-Memory~\cite{gmemory} organize memory as temporal or knowledge graphs but without granularity separation.
Unlike these works, we factor memory into instance-grounded exemplars and category-level IF-THEN schemas synthesized exclusively from a transient reflection store, operationalizing schema-mediated consolidation~\citep{tse2007schemas} so that exemplar text never contaminates schema synthesis.

\paragraph{Construction Regime.} Memory-R1~\cite{memory-r1}, Mem-$\alpha$~\cite{mem-alpha}, Memory-as-Action~\cite{memoryasaction}, and MEM1~\cite{mem1} learn ``add/update/delete'' policies under outcome rewards. MemBuilder~\cite{membuilder} trains a lightweight language model as the memory manager with attributed dense rewards; MemGen~\cite{memgen} fuses RL-generated latent memory tokens into the decoder; Mem-T~\cite{mem-t} and LatentMem~\cite{latentmem} apply GRPO over Memory-Operation Trees and learnable composers respectively.
These methods require gradients, often opaque latent representations, and a fixed cognitive taxonomy.
A complementary line distills memory offline from rollouts of a frozen backbone~\cite{expel, awm, reasoningbank, vilomem, ace}; we belong here.
First, our online concept categorizer grows the category space incrementally during the training pass instead of committing to a predefined taxonomy, comparable in spirit to STITCH's contextual-intent indexing~\cite{stitch} but operating over reasoning concepts rather than dialogue intents.
Second, we add an inter-test-time attribution pass (Stage~2) that approximates RL-style credit assignment without ever updating model parameters, slotting between similarity-only distillation (AWM~\cite{awm}, ReasoningBank~\cite{reasoningbank}) and full RL of memory managers (Memory-R1~\cite{memory-r1}, MemGen~\cite{memgen}).

\begin{table}[t]
\caption{\textbf{Comparison of agentic memory methods across five design dimensions}: whether the method (i)~handles \emph{multimodal} inputs (vision+text), (ii)~maintains \emph{instance-grounded exemplars}, (iii)~maintains \emph{abstract schema rules}, (iv)~performs \emph{per-rule credit assignment} beyond similarity-only retrieval, and (v)~is \emph{gradient-free} (no model parameter updates). \cmark\,/\,\xmark\ indicates whether the method satisfies a given dimension. See the surrounding text for column definitions.}
\label{tab:related-comparison}
\centering
\footnotesize
\setlength{\tabcolsep}{4pt}
\begin{tabular}{l|ccccc}
\toprule \toprule
Method & Multimodal & Exemplar & Schema & Credit Assn. & Grad-Free \\
\midrule
Mem0~\cite{mem0}                    & \xmark & \cmark & \xmark & \xmark & \cmark \\
Memento~\cite{memento}              & \xmark & \cmark & \xmark & \cmark & \cmark \\
A-Mem~\cite{amem}                   & \xmark & \cmark & \xmark & \xmark & \cmark \\
ExpeL~\cite{expel}                  & \xmark & \xmark & \cmark & \xmark & \cmark \\
AWM~\cite{awm}                      & \xmark & \xmark & \cmark & \xmark & \cmark \\
ReasoningBank~\cite{reasoningbank}  & \xmark & \xmark & \cmark & \xmark & \cmark \\
ACE~\cite{ace}                      & \xmark & \xmark & \cmark & \xmark & \cmark \\
ViLoMem~\cite{vilomem}              & \cmark & \xmark & \cmark & \xmark & \cmark \\
TFGRPO~\cite{tfgrpo}                & \xmark & \xmark & \cmark & \cmark & \cmark \\
D2Skill~\cite{d2skill}              & \xmark & \cmark & \cmark & \xmark & \cmark \\
Memory-R1~\cite{memory-r1}          & \xmark & \cmark & \xmark & \cmark & \xmark \\
MemGen~\cite{memgen}                & \xmark & \xmark & \xmark & \cmark & \xmark \\
\midrule
\textbf{DG-Mem (Ours)}             & \cmark & \cmark & \cmark & \cmark & \cmark \\
\bottomrule \bottomrule
\end{tabular}
\vspace{-2mm}
\end{table}

\section{Methods}
\label{sec:methods}

\begin{figure}[htbp]
    \centering
    \begin{subfigure}[b]{0.54\textwidth}
        \centering
        \includegraphics[width=\textwidth]{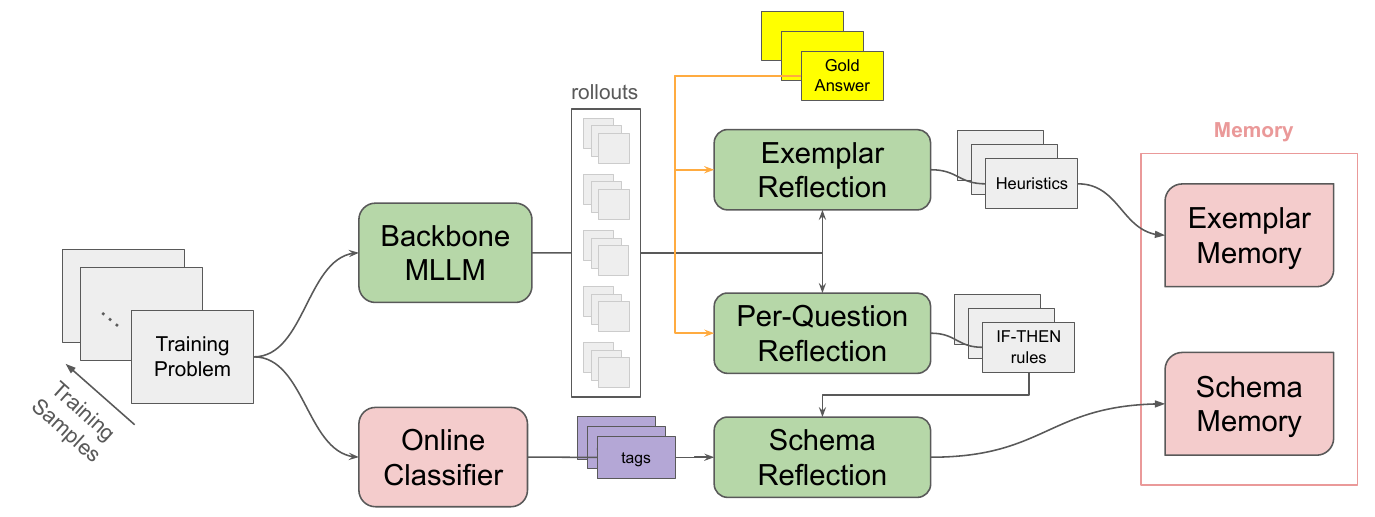}
        \label{fig:stage1}
    \end{subfigure}
    \hfill \vrule width 0.5pt \hfill
    \begin{subfigure}[b]{0.44\textwidth}
        \centering
        \includegraphics[width=\textwidth]{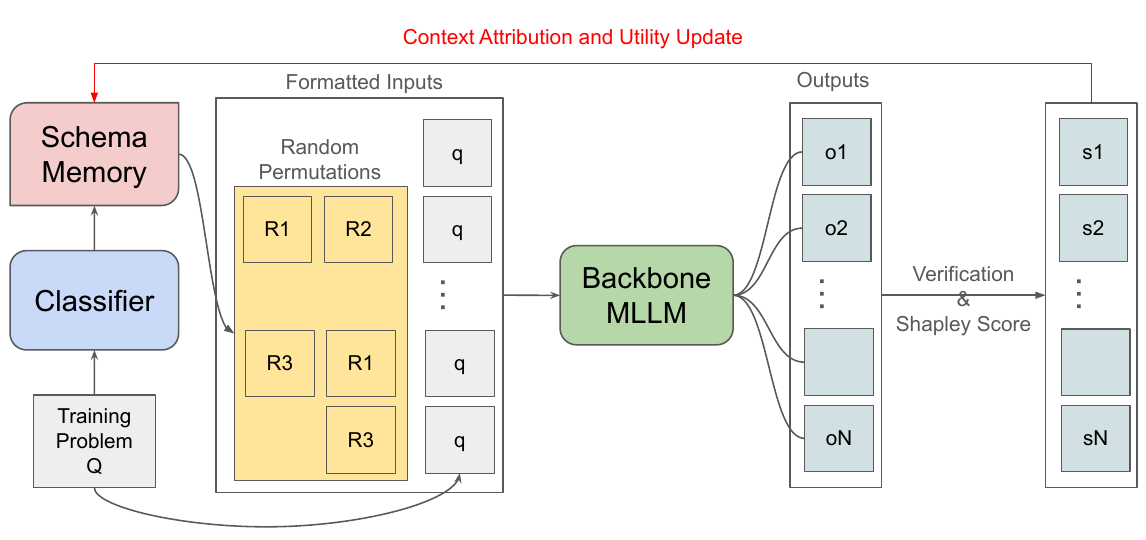}
        \label{fig:stage2}
    \end{subfigure}

    \caption{\textbf{Two-stage offline pipeline for memory training on a frozen MLLM backbone.}
    \textbf{Left (Stage 1 -- Memory Construction):} for each training problem $q$, $\MLLM$ produces $N$ rollouts that an online classifier and three parallel reflectors turn into an exemplar memory $\calM_E$ and a category-level schema memory $\calM_S$, with a transient reflection store $\calM_R$ mediating the latter so that schema synthesis never reads exemplar text.
    \textbf{Right (Stage 2 -- Shapley Context Attribution):} the same frozen $\MLLM$ is reused as an attribution device; for each problem we draw $B$ subsets of the active category's rules, run a rollout per subset, and convert the binary rewards into per-rule Shapley contributions that are accumulated into a utility column on $\calM_S$ and used to re-weight retrieval at test time.
    See \S\ref{sec:methods:construction} and \S\ref{sec:methods:utility} for full details.}
    \vspace{-4mm}
\end{figure}

\paragraph{Cognitive Foundations.} The introduction motivates DG-Mem from CLS~\citep{mcclelland1995complementary, kumaran2016learning}; here we summarize the operational consequences. We instantiate the fast, instance-specific store as an exemplar memory $\calM_E$~\citep{nosofsky1986attention} and the slow, structured store as a schema memory $\calM_S$, with an intermediate reflection store $\calM_R$ acting as the unconsolidated episodic trace from which schemas are later synthesized~\citep{tse2007schemas}. At inference, $\calM_E$ and $\calM_S$ are accessed through two complementary mechanisms, \emph{priming} (top-down context before reasoning) and \emph{verification} (post-hoc audit). Following the rational analysis of memory~\citep{anderson1991reflections} and evidence that offline replay is biased toward rewarded experience~\citep{singer2009rewarded,shohamy2010dopamine}, retrieval composes embedding similarity with a per-rule utility learned offline, so that two memories with identical cue match are tied by their historical track record. We treat these correspondences as design intuitions rather than cognitive claims.

\paragraph{Agentic Architecture.}
We augment a frozen MLLM backbone $\MLLM$ with a non-parametric, externally stored memory $\calM$ that is constructed once from a training corpus and consulted read-only at test time.
No gradients propagate into $\MLLM$ or additional layers.
For each test problem $q$, the agent executes a short, memory-aware decision pipeline:
it analyzes $q$ to produce a concept summary $\calK(q)$,
queries $\calM$ with $\calK(q)$ and the multimodal representation $(I_q, t_q)$ to obtain a relevant subset $\calM_q \subseteq \calM$,
and conditions the backbone on the composed context to produce an answer $\hat{a}_q = \MLLM(q, \calM_q)$.
The remainder of this section specifies how $\calM$ is constructed (\S\ref{sec:methods:construction}), how it is accessed (\S\ref{sec:methods:access}), and how utility estimates refine what is retained (\S\ref{sec:methods:utility}).

\subsection{Memory Construction}
\label{sec:methods:construction}

Memory is built offline from rollouts of $\MLLM$ on a training corpus.
For each problem $q$ with gold answer $a_q^\star$, we draw $N$ temperature-sampled solution trajectories $\{\MLLM_i(q)\}_{i=1}^{N}$ and label each \textsc{correct} or \textsc{incorrect} via the verification cascade $\Verify(\MLLM_i(q), a_q^\star)\in\{0,1\}$ (native benchmark matcher, symbolic parser, then LLM judge).
The labeled rollout group is the only training signal; we do not require reasoning annotations beyond $a_q^\star$.

Given the labeled rollout group, three summarizations are produced in parallel and written to distinct stores:
\begin{align}
\rho(q) &= \Refl_R\bigl(q,\, a_q^\star,\, \{\MLLM_i(q)\}_{i=1}^{N}\bigr) &&\text{(IF-THEN rules)}, \label{eq:refl-r}\\
e(q) &= \Refl_E\bigl(q,\, a_q^\star,\, \{\MLLM_i(q)\}_{i=1}^{N}\bigr) &&\text{(exemplar record)}, \label{eq:refl-e}\\
\kappa(q) &= \Cat(q) &&\text{(category tag)}. \label{eq:cat}
\end{align}
The classifier $\Cat$ is the only model that consumes $q$ alone: it routes the problem against the existing category registry without inspecting the rollouts.
The three stores these artifacts feed into are related by a strict dependency.
The reflection store $\calM_R = \{\rho(q)\}_q$ is an intermediate object never retrieved at inference; it exists solely as source material for the two downstream stores.
The exemplar memory $\calM_E$ co-locates $\rho(q)$ inside each exemplar, so that a single retrieval delivers both the instance-grounded strategy and its associated abstract rules.
The schema memory $\calM_S$ is synthesized after the pass completes, \emph{exclusively} from $\calM_R$ grouped by $\kappa$, without ever consulting the contents of $\calM_E$.

All three summarizations are bound by an \emph{abstraction constraint}: numeric values, option letters, variable bindings, and named entities drawn from $q$ are forbidden in any stored rule, strategy, or check.
This is a correctness requirement, not a stylistic preference, because each stored item is consumed in the context of a different problem at test time; any leaked specifics would propagate directly into the solver's conditioning.

\paragraph{Exemplar Memory.}
An exemplar $e(q) \in \calM_E$ is a self-contained record engineered to transfer a single training problem's lesson to a new, conceptually similar problem.
The reflector $\Refl_E$ distills the labeled rollout group into a two-field summary: a \emph{reasoning strategy} that verbalizes a step-by-step procedure applicable to any problem of the same type, and a \emph{verification check} that names the dominant error mode exhibited by the incorrect trajectories.

Each exemplar is indexed by a multimodal embedding $\Emb_M(I_q, t_q)$, where $I_q$ is the problem image and $t_q$ is the short query \textit{``This problem tests $\calK(q)$. Question: $q$.''}.
Joint indexing matters in practice: problems that share textual concepts but differ visually -- e.g., a full circle vs.\ a circular sector, or a bar chart vs.\ a stacked bar chart -- are near-miss distractors that a purely textual retriever cannot separate.

\paragraph{Schema Memory.}
A schema $\sigma_\kappa \in \calM_S$ captures what is stable across a concept-equivalent class of problems.
Once the online pass over the training set has assigned every problem a category $\kappa(q)$, we partition $\calM_R$ by category and apply the \emph{schema synthesizer} $\Synth$ to each partition:
\begin{equation}
\label{eq:schema-synth}
\sigma_\kappa = \Synth\!\left(\kappa,\; \concepts(\kappa),\; \bigcup_{q:\,\kappa(q)=\kappa} \rho(q)\right),
\end{equation}
where $\concepts(\kappa)$ denotes the representative key concepts attached to category $\kappa$ in the registry.
$\Synth$ is instructed to eliminate redundancy, reconcile near-duplicate rules, and elevate instance-specific phrasing into principles applicable to the entire category.
Its output is a compact list of prescriptive advisories that tell the solver what to do rather than merely what to avoid.
Two properties of this construction warrant emphasis. First, consistent with the schema-mediated consolidation view~\citep{tse2007schemas}, synthesis operates strictly over $\calM_R$: trajectories and exemplar-level text are never shown to $\Synth$, confining $\sigma_\kappa$ to principles derivable from abstract rules and limiting the risk of memorizing solution strings. Second, because $\sigma_\kappa$ aggregates over many problems sharing $\kappa$, a spurious rule from any single $\Refl_R$ call is diluted rather than amplified, endowing $\calM_S$ with a robustness property that $\calM_E$, being instance-grounded, does not share.

\paragraph{Online Categorization.}
Partitioning $\calM_R$ requires a category label per problem, but a fixed taxonomy is either too coarse (a single ``geometry'' bucket) or too fine (one bucket per concept signature), and the right granularity is unknown a priori.
We instead grow the registry online: $\Cat$ extracts $\calK(q)$ and compares it against the running registry restricted to identifier-name pairs (to minimize prompt overhead), then either assigns $q$ to an existing category or introduces a new one with a fresh identifier and representative concepts.
Registry updates are serialized so the category space grows monotonically; the resulting $\kappa(q)$ is recorded alongside $\rho(q)$ and $e(q)$, and the classification step runs concurrently with $\Refl_E$ and $\Refl_R$ at no sequential overhead.

\subsection{Memory Access}
\label{sec:methods:access}

At test time $\calM=\calM_S\cup\calM_E$ is frozen; $\calM_R$ is never consulted directly, since its content reaches inference only through the two derived stores.
Access proceeds in three retrieval stages, followed by a fallback mechanism against reasoning-loop collapse.

\paragraph{Schema Retrieval.}
Schemas are retrieved at the granularity of individual rules within the categories whose representative concepts most closely match $\calK(q)$.
Cue similarity between rule $r_i$ and the query is computed as $\operatorname{sim}(r_i, q) = \cos\bigl(\Emb_T(r_i),\, \Emb_T(q)\bigr)$ using the text embedder $\Emb_T$.
We compose cue similarity with the rule's accumulated utility $U(r_i)$ (defined in \S\ref{sec:methods:utility}):
\begin{equation}
\label{eq:schema-score}
    s_i = \alpha\,\widehat{\operatorname{sim}}(r_i, q)+(1-\alpha)\,\widehat{U}(r_i),
\end{equation}
where $\widehat{(\cdot)}$ denotes min-max normalization across the candidate set under the active category, applied independently to similarity and utility so that magnitude differences between the two scales do not dominate the ranking.
$\alpha=1$ collapses to similarity-only retrieval and recovers a memory-only baseline; $\alpha=0$ ranks purely by accumulated utility.

\paragraph{Exemplar Retrieval.}
Exemplars are retrieved by cosine similarity between the multimodal embedding of the query $\Emb_M(I_q, t_q)$ and the stored multimodal embedding $\Emb_M(I_{q'}, t_{q'})$ of each candidate exemplar $e(q')\in\calM_E$.
A similarity threshold $\tau_E$ filters the candidate pool: if no candidate clears $\tau_E$, no exemplar is injected.

\paragraph{Memory Injection.}
Retrieved memories are inserted into the solver's context in one of two complementary modes: \emph{priming}, which precedes the question and frames the solver's approach a priori, and \emph{verification}, which follows the question and reserves the same content for a post-hoc audit of the produced answer.

\subsection{Utility Training}
\label{sec:methods:utility}

After Stage 1 every rule in $\calM_S$ is treated identically by retrieval, yet a category's rules are rarely uniformly useful: some encode dominant failure modes that transfer broadly, while others are narrow tips whose conditions seldom fire. 
A second offline pass turns the same backbone into an attribution device: it draws additional rollouts in which the only varying factor is which subset of category rules is injected, and converts the resulting reward signal into a per-rule utility column that re-weights retrieval at test time.

\paragraph{Frozen-Category Assignment.} 
Stage 2 reads back the schema memory $\calM_S$ and the category registry produced in Stage 1; both are frozen for the remainder of training so that utility statistics accumulate on a fixed rule support.

\paragraph{Subset Rollouts.}
For problem $q$ we draw a batch of $B$ rule subsets $\{S_b\}_{b=1}^{B}$ that always contains the empty subset $\emptyset$ (query-only baseline) and the full subset $\calR_{\kappa^\star}$; the remaining $B-2$ subsets are sampled uniformly without replacement from the strict middle layers $\{S\subseteq\calR_{\kappa^\star}:1\le|S|\le K-1\}$.
When $B\ge 2^K$ the batch is the exact powerset and Shapley becomes exact.
For each subset, the backbone is conditioned on the same problem with only the injected rule set varying, $o_b = \MLLM(q, S_b)$, and the resulting binary reward $v(S_b) = \Verify\bigl(o_b,\, a_q^\star\bigr) \in \{0,1\}$
isolates the contribution of $S_b$ itself, since temperature, decoding seed, and image content are held fixed across $b$.

\paragraph{Context Attribution.}
Treating rules as players and the observed subset rewards $v(\cdot)$ as the cooperative value, the contribution of rule $r_i$ to problem $q$ is its average marginal gain across all permutations of $\calR_{\kappa^\star}$,
\begin{equation}
\label{eq:shapley}
\Delta_q(r_i) = \frac{1}{K!}\sum_{\pi\in\mathfrak{S}_{K}}\!\Bigl[v\bigl(P^\pi_{i}\cup\{r_i\}\bigr) - v\bigl(P^\pi_i\bigr)\Bigr],
\end{equation}
where $P^\pi_i$ is the set of rules preceding $r_i$ in $\pi$ and unobserved coalitions are imputed as $0$. Shapley distributes credit unevenly among rules in proportion to where each one most increases the marginal reward, at an $O(K!)$ cost in the exact limit.

Each rule $r_i$ maintains a running sum and a count $n_i$ of how many problems' batches updated it; its accumulated utility is the empirical mean $U(r_i) = {\sum_{q\,:\,r_i\in\calR_{\kappa(q)}} \Delta_q(r_i)} / {\max(n_i, 1)}$.
Sums and counts are merged into the persisted category file under file-level locking, so that Stage 2 can be parallelized across problems and resumed without double-counting.
Exemplars are intentionally excluded from this update: because Stage-2 problems can collide with Stage-1 problems, attributing utility to an exemplar whose embedding indexes the same problem would leak.
Exemplars therefore continue to be ranked by similarity alone.

At test time, $U(r_i)$ is plugged into Eq.~\ref{eq:schema-score}, completing the rational-analysis composition: two rules with equal cue match are ranked by their accumulated track record.

\section{Experiments}
\label{sec:experiments}

\subsection{Main Results on Multimodal Benchmarks}

\paragraph{Tasks and datasets.} 
Our target benchmarks are mainly multi-modal scientific problem solving and reasoning, including MathVista~\cite{mathvista}, MMMU~\cite{mmmu}, MMMU-Pro~\cite{mmmu-pro}. 
For each benchmark we adopt a fixed 3:1 train/test split, where the test set is a subset of the full benchmark. Only the gold answer of training problems are needed as supervision; no chain-of-thought or per-step reasoning annotations are required.
The training partition is consumed exclusively offline to construct $\calM_E$ and $\calM_S$, and the test partition is held out for all reported numbers.
At evaluation time the memory store is frozen: no rollouts, reflections, or schema updates are produced during testing.

\paragraph{Models} 
We instantiate DG-Mem with four backbones including open-weight and proprietary regimes: Qwen3.5-27B~\cite{qwen35}, Qwen3.5-122B-A10B~\cite{qwen35}, GPT-5-Nano~\cite{gpt-5}, and Gemini-3-Flash~\cite{gemini-3-flash}.
Except for the LLM judge used by the verification cascade, which is fixed to Qwen-3.5-Flash~\cite{qwen35} across all experiments, every node in the agent graph (rollout solver, reflectors, schema synthesizer, online categorizer) uses the \emph{same backbone}.
This homogeneous configuration enforces the \emph{``self-evolving''} setting and prevents implicit knowledge distillation from a larger or heterogeneous model.
All text retrievers use Qwen-3-Embedding~\cite{qwen3-embedding}; all multimodal retrievers use Qwen-3-VL-Embedding~\cite{qwen3-vl-embedding}.

\paragraph{Baselines.}
We compare against two families of methods that share our no-additional-supervision constraint.
\textbf{ViLoMem}~\cite{vilomem} is a training-free dual-stream memory agent that grows logical and visual error guidelines online during evaluation; it represents the ``memory without offline consolidation'' regime.
We implement ViLoMem into two settings: offline and online. For offline, it freezes the trained memory and evaluates on the test split. For online, it continues to grow and refine the memory with the test split. We report the top-1 accuracies on the test split.
We additionally report a \textbf{No-memory} baseline that runs the same backbone with an identical decoding configuration but no retrieval, isolating the contribution of the memory pipeline itself.

\paragraph{Evaluation metrics.} We report top-1 accuracy on the held-out test split of each benchmark, judged by the same verification cascade used during memory construction (native matcher $\to$ symbolic parser $\to$ LLM judge); reporting the cascade's verdict avoids spurious gaps caused by formatting differences.

Table~\ref{tab:main-results} reports test accuracy across the three benchmarks for each backbone, comparing the No-memory baseline, ViLoMem~\cite{vilomem}, and DG-Mem.
DG-Mem delivers consistent gains over the No-memory baseline across backbones.

\begin{table}[h]
    \vspace{-3mm}
    \caption{\textbf{Top-1 accuracy (\%) on held-out test splits across multimodal reasoning benchmarks.} Each backbone is evaluated under four conditions: \emph{No-memory} (same backbone and decoding configuration with retrieval disabled), \emph{ViLoMem (offline)} (logical/visual error memory frozen after training), \emph{ViLoMem (online)} (memory continues to grow on the test split), and \emph{DG-Mem} (Ours). \emph{Avg Imp.} reports the per-backbone mean improvement over the No-memory baseline across the three benchmarks. * on each benchmark name indicates that we report on a fixed held-out test partition rather than the full benchmark. Best result per backbone is in \textbf{bold}.}
    \centering
    \begin{tabular}{ll|ccc|c}
    \toprule \toprule
        Backbone & Method & MathVista* & MMMU* & MMMU-Pro* & Avg Imp. \\
    \midrule
        \multirow{4}{*}{Qwen3.5-122B-A10B}
            & No-memory & 83.6 & 85.8 & 77.9 & - \\
            & ViLoMem (offline) & 85.2 & 86.4 & 80.0 & +1.4 \\
            & ViLoMem (online) & 88.4 & 87.4 & 80.3 & +2.9 \\
            & \textbf{DG-Mem (Ours)} & \textbf{88.7} & \textbf{89.2} & \textbf{86.1} & +5.6 \\
    \midrule
        \multirow{4}{*}{Qwen3.5-27B}
            & No-memory & 82.9 & 83.4 & \textbf{75.1} & - \\
            & ViLoMem (offline) & 86.4 & 85.9 & 74.6 & +1.8 \\
            & ViLoMem (online) & \textbf{87.2} & 86.3 & 74.8 & +2.3 \\
            & \textbf{DG-Mem (Ours)} & 87.0 & \textbf{87.9} & 73.6 & +2.3 \\
    \midrule
        \multirow{4}{*}{GPT-5-Nano}
            & No-memory & 64.8 & 67.6 & 48.2 & - \\
            & ViLoMem (offline) & 66.8 & 56.9 & 51.0 & -2.0 \\
            & ViLoMem (online) & 70.4 & 58.0 & 52.5 & +0.1 \\
            & \textbf{DG-Mem (Ours)} & \textbf{74.4} & \textbf{79.4} & \textbf{64.3} & +12.5 \\
    \midrule
        \multirow{4}{*}{Gemini-3-Flash}
            & No-memory & 87.2 & 82.4 & 69.0 & - \\
            & ViLoMem (offline) & 80.4 & 82.8 & 73.8 & -0.2 \\
            & ViLoMem (online) & 83.6 & 82.4 & 71.1 & -0.2 \\
            & \textbf{DG-Mem (Ours)} & \textbf{91.6} & \textbf{84.7} & \textbf{77.1} & +5.2 \\
    \bottomrule \bottomrule
    \end{tabular}
    \label{tab:main-results}
\end{table}




\subsection{Ablation Studies}

We isolate the contribution of each component by ablating one piece of the pipeline at a time on the Gemini-3-Flash backbone.
Table~\ref{tab:ablation} summarizes the sweep; the paragraphs below interpret each row.

\begin{table}[h]
    \caption{Component ablations on Gemini-3-Flash. Each row disables a single mechanism while leaving the rest of the pipeline intact. 
    \emph{w/o $\calM_E$} retains the category-level schema memory but removes the exemplar store, so retrieval returns only IF-THEN rules.
    \emph{w/o $\calM_S$} keeps exemplars but disables schema retrieval, so the prompt is augmented only with nearest-neighbor solved instances.
    \emph{w/o utility} sets the retrieval interpolation $\alpha{=}1$, ranking rules by embedding similarity alone and discarding the Stage-2 Shapley utility column.
    Colored deltas (\green{green}/\red{red}) report the absolute change in accuracy relative to the No-memory \emph{Baseline} row.}
    \centering
    \begin{tabular}{l|ccc}
    \toprule \toprule
        Datasets & MathVista* & MMMU* & MMMU-Pro* \\
    \midrule
\; Baseline & 86.2 & 82.4 & 69.0 \\
    \midrule
\;-- w/o $\calM_E$ (stage-1 schema only) & 88.4 \green{+2.2} & 82.1 \red{-0.3} & 73.4 \green{+4.4} \\
\;-- w/o $\calM_S$ (stage-1 exemplar only) & 87.6 \green{+1.4} & 84.0 \green{+1.6} & 73.4 \green{+4.4} \\
\;-- w/o utility ($\alpha{=}1$, stage-1 both memory) & 91.0 \green{+4.8} & 84.0 \green{+1.6} & 75.0 \green{+6.0} \\
    \midrule
        Full method & 91.6 \green{+5.4} & 84.7 \green{+2.3} & 77.1 \green{+8.1} \\
    \bottomrule \bottomrule
    \end{tabular}
    \label{tab:ablation}
    \vspace{-2mm}
\end{table}

\paragraph{Effects of Exemplar Memory.}
Removing $\calM_E$ from the full method costs $-3.2$, $-2.6$, and $-3.7$ on MathVista/MMMU/MMMU-Pro, with the largest drop on MMMU-Pro where many problems hinge on a specific diagrammatic configuration (circuit topologies, geometric constructions) that an IF-THEN rule abstracts away but a multimodal nearest neighbor preserves.
Schema-only is also the only condition that falls slightly below the No-memory baseline on MMMU ($82.1$ vs.\ $82.4$), indicating that schemas alone can over-generalize on broad multi-discipline questions and that $\calM_E$ acts as a corrective check against schema misfires.

\vspace{-2mm}
\paragraph{Effects of Schema Memory.}
The exemplar-only ablation trails the full method by $-4.0$, $-0.7$, and $-3.7$, with the largest gap on MathVista where many test problems share a high-level solution template (parsing a bar chart, applying a unit conversion) but are visually dissimilar from any single training instance -- exactly the regime where a transferable rule beats a literal nearest neighbor. That exemplar-only and schema-only land within $\sim 1\%$ of each other yet the full method exceeds both indicates the two grains cover non-overlapping failure modes rather than redundant signal, which is what the schema / exemplar separation through $\calM_R$ is designed to enforce.

\vspace{-2mm}
\paragraph{Effects of Utility.}
Setting $\alpha{=}1$ (similarity-only retrieval over $\calM_S$) costs $-0.6$, $-0.7$, and $-2.1$ relative to the full method, with the gap widening on the harder MMMU-Pro split.
This shows that not every embedding-similar rule is useful: many are too generic, fitted to a different sub-population of the category, or actively misleading; Shapley-derived utility re-weights retrieval toward historically reward-bearing rules.
The absolute margin is small because similarity already narrows the candidate set, but the consistent direction across all three benchmarks indicates utility is doing real work rather than tie-breaking, and we expect it to widen as $\calM_S$ accumulates more superficially similar rules per category.

\subsection{Qualitative Results}

\paragraph{Qualitative cases where dual-grained memory corrects the backbone.}
Figure~\ref{tab:qualitative} shows three test instances on which the No-memory backbone is wrong but the same backbone with our retrieved memory is correct, and the same correction also occurs under category-only and exemplar-only retrieval (i.e., the fix is not specific to one grain).
For each case we surface the retrieved exemplar (instance-grounded $\calM_E$) and the most relevant schema rule (category-level $\calM_S$) so that the source of the correction is auditable rather than incidental.
The three cases span visual enumeration after attribute filtering (CLEVR-style subtraction), threshold counting from a chart, and entity-grounded arithmetic over public figures, and in each case the retrieved schema names the procedure the baseline omitted while the retrieved exemplar supplies a near-isomorphic solved template.

\begin{figure}[t]
\centering
\tiny
\renewcommand{\arraystretch}{1.0}
\setlength{\tabcolsep}{2pt}
\begin{tabular}{@{}p{0.20\linewidth}p{0.27\linewidth}p{0.27\linewidth}p{0.18\linewidth}@{}}
\toprule
\textbf{Test question} & \textbf{Retrieved exemplar ($\calM_E$)} & \textbf{Retrieved schema rule ($\calM_S$)} & \textbf{Baseline $\to$ Ours} \\
\midrule
\emph{MathVista \#90.}
``Subtract all yellow matte blocks. Subtract all tiny brown cylinders. How many objects are left?''
&
Solved template: ``Subtract all red things. Subtract all tiny matte balls. How many objects.''
\emph{Strategy:} ``Catalog every object's color, size, material, and shape, then translate the natural-language descriptors to those attribute categories before subtracting.''
&
\emph{Visual Counting and Attribute Analysis:} ``Perform an exhaustive initial census of all objects, assign a count of zero to any missing categories, and calculate the cardinality of each subset \emph{independently} before applying the subtraction.''
&
\textbf{Gold:} 5 \newline
\red{Baseline: 4} (skipped enumeration; ``There are 4 objects left.'') \newline
\green{Ours: 5} (lists 7 objects, removes 2) \\
\midrule
\emph{MathVista \#13.}
``How many objects are preferred by more than 90\% of people in at least one category?'' (grouped bar chart)
&
Solved template: ``How many algorithms have accuracy higher than 9 in at least one dataset?''
\emph{Verification check:} ``Re-evaluate any bar that appears to end exactly on a midpoint or grid line to confirm whether it meets a `strictly greater than' or `strictly less than' condition.''
&
\emph{Data Analysis and Interpretation:} ``When identifying threshold-based counts, \emph{systematically} compute the metric for every distinct item rather than relying on visual estimation of distances or assumed ordering.''
&
\textbf{Gold:} 0 \newline
\red{Baseline: 2} (visually estimated bars near the 90\% line) \newline
\green{Ours: 0} (computes per-bar height, applies strict $>$) \\
\midrule
\emph{MathVista \#792.}
``What is the age gap between these two people?'' (photo of P.\ Hammond and M.\ Kaljurand)
&
Solved template: same age-gap question structure on a different pair.
\emph{Verification check:} ``Cross-verify the identity of the \emph{less prominent} individual against the primary subject's known associates or event context, as misidentification of this figure is the most frequent source of error.''
&
\emph{Arithmetic: Age Problems and Data Extraction:} ``Subtract the earlier birth year from the later birth year, ensuring the values belong to the \emph{exact subjects specified} in the query.''
&
\textbf{Gold:} 7 \newline
\red{Baseline: 3} (used wrong birth year for Kaljurand: 1958) \newline
\green{Ours: 7} (corrects to 1962, then $1962{-}1955$) \\
\bottomrule
\end{tabular}
\caption{\textbf{Qualitative cases where DG-Mem corrects the No-memory backbone.}
Each row is a held-out test instance for which both retrieved grains are populated and the same baseline error is corrected under \emph{category-only}, \emph{exemplar-only}, and \emph{both-memory} retrieval, ruling out the alternative explanation that any prompt perturbation flips the answer.
The retrieved schema rule names the procedure the baseline omitted (an exhaustive census; per-item threshold computation; grounding the arithmetic in the correct entities), and the retrieved exemplar supplies a solved template whose verification check directly addresses the failure mode (boundary-bar evaluation; identity disambiguation of the less prominent subject).}
\label{tab:qualitative}
\vspace{-2mm}
\end{figure}

\section{Limitations}
\label{sec:limitations}

We highlight several limitations that may bound the scope of our claims and point to natural directions for follow-up work.

\paragraph{Stochasticity of attribution.}
Shapley context attribution treats the backbone as a deterministic value function on rule subsets. 
But non-deterministic GPU kernels, dynamic batching, and floating-point non-associativity~\cite{he2025nondeterminism} mean binary rewards $v(S_b)$ can vary across identical-context calls.
With a finite subset budget $B$, $\Delta_q(r_i)$ inherits this background noise: a rule with small true marginal effect can register non-zero $\Delta_q$ from answer-flipping unrelated to the rules injected, and the sign of $U(r_i)$ becomes data-order-dependent.
We mitigate by averaging across many problems sharing the same rule and ranking rather than thresholding on $U(r_i)$, but the residual variance is a property of the value function; multi-rollout majority-vote rewards would harden this further at additional training cost.

\paragraph{Frozen memory at test time.}
We freeze $\calM$ at evaluation time so that test accuracy reflects retrieval and conditioning rather than additional online learning, and so that comparisons against ViLoMem (offline) are head-to-head.
This is a clean experimental choice but forfeits the continual-learning regime in which test-time errors could themselves seed new exemplars or rules.
Combining offline consolidation with bounded online updates remains an open design question.

\section{Conclusions}
\label{sec:conclusions}

We presented \textbf{DG-Mem}, a dual-grained agentic memory framework that equips frozen multimodal backbones with an instance-grounded exemplar memory and a category-level schema memory, strictly separated by a transient reflection store. DG-Mem addresses two key challenges in agentic memory. 
First, online incremental categorization dynamically tracks the corpus's empirical concept distribution without a fixed, manual taxonomy. 
Second, Shapley context attribution repurposes the frozen backbone for credit assignment, optimizing retrieval ranking by empirical utility rather than mere similarity. 
This gradient-free pipeline is readily deployable on both closed-weight and on-device models.

Empirically, DG-Mem consistently outperforms memory-less baselines across four backbones and three multimodal reasoning benchmarks, as well as a strong single-grain baseline (ViLoMem) in competitive configurations. Ablations confirm that exemplar and schema memories address distinct failure modes, while utility re-weighting yields a consistent margin expected to widen as the memory scales. Ultimately, DG-Mem operationalizes a Complementary Learning Systems view of agentic memory, establishing attribution estimators and bounded online consolidation as highly promising directions for future research.

\clearpage
\bibliographystyle{plain}
\bibliography{references}

\clearpage
\appendix
\section*{Appendix}

\section{Additional Experimental Details}

\subsection{Datasets and Splits}
\label{sec:appendix:exp-details:splits}

We evaluate on three multimodal reasoning benchmarks and use a fixed 3:1 train/test partition per benchmark.
Only the gold answer of training problems is consumed as supervision; no chain-of-thought or step-level annotations are used.
Train problems drive Stage-1 memory construction and Stage-2 utility attribution; test problems are held out and evaluated with the memory frozen.

\begin{table}[h]
\caption{Train/test partitions per benchmark. The same split file is used by every backbone and by every method (No-memory, ViLoMem, DG-Mem) so that comparisons are head-to-head.}
\label{tab:appendix:splits}
\centering
\small
\begin{tabular}{l|cc|c}
\toprule \toprule
Benchmark & Train (Stage 1 \& 2) & Test (held out) & Total \\
\midrule
MathVista (MINI)  & 750  & 250 & 1{,}000 \\
MMMU (DEV+VAL)    & 788  & 262 & 1{,}050 \\
MMMU-Pro          & 1{,}298 & 432 & 1{,}730 \\
\bottomrule \bottomrule
\end{tabular}
\end{table}

\subsection{Test-Time Retrieval}
\label{sec:appendix:exp-details:test}

At test time $\calM = \calM_S \cup \calM_E$ is frozen.
For each query we run the concept summarizer $\calK$ once, retrieve the top categories whose representative concepts clear the category-similarity gate, then retrieve up to $K_{\max}$ rules per active category ranked by Eq.~\eqref{eq:schema-score}.
Exemplars are retrieved by multimodal cosine similarity against $\Emb_M(I_q,t_q)$ subject to a similarity-threshold gate; if no exemplar clears the gate, none is injected.

\begin{table}[h]
\caption{Test-time retrieval and decoding hyperparameters.}
\label{tab:appendix:test-hp}
\centering
\small
\begin{tabular}{l|c|l}
\toprule \toprule
Hyperparameter & Value & Description \\
\midrule
$\alpha$                       & 0.8    & Similarity / utility interpolation in Eq.~\eqref{eq:schema-score} \\
Category similarity threshold  & 0.40   & Min cosine similarity for category retrieval \\
Schema rule similarity threshold & 0.35 & Min cosine similarity for rule retrieval \\
$K_{\max}$ (rules per category) & 2     & Top-$k$ rules retrieved per active category \\
\bottomrule \bottomrule
\end{tabular}
\end{table}

\section{Additional Ablation Studies}

We complement the Gemini-3-Flash ablations in the main text (Table~\ref{tab:ablation}) with the same component sweep on two additional backbones, GPT-5-Nano (Table~\ref{tab:ablation-gpt}) and Qwen3.5-122B-A10B (Table~\ref{tab:ablation-qwen}). 
The pattern is consistent across both backbones: each grain alone improves over the No-memory baseline on every benchmark, and the full method exceeds the better of the two single-grain ablations on every cell, confirming that exemplar and schema memories cover non-overlapping failure modes rather than redundant signal.

\begin{table}[h]
    \caption{Component ablations on GPT-5-Nano. Each row disables a single mechanism while leaving the rest of the pipeline intact.}
    \centering
    \begin{tabular}{l|ccc}
    \toprule \toprule
        Datasets & MathVista* & MMMU* & MMMU-Pro* \\
    \midrule
\; Baseline & 64.8 & 67.6 & 48.2 \\
    \midrule
\;-- w/o $\calM_S$ (stage-1 exemplar only) & 68.8 \green{+4.0} & 72.5 \green{+4.9} & 55.6 \green{+7.4} \\
\;-- w/o $\calM_E$ (stage-2 schema only) & 67.2 \green{+2.4} & 74.0 \green{+6.4} & 58.1 \green{+9.9} \\
    \midrule
        Full method & 74.4 \green{+9.6} & 79.4 \green{+11.8} & 64.4 \green{+16.2} \\
    \bottomrule \bottomrule
    \end{tabular}
    \label{tab:ablation-gpt}
\end{table}

\begin{table}[h]
    \caption{Component ablations on Qwen3.5-122B-A10B. Each row disables a single mechanism while leaving the rest of the pipeline intact.}
    \centering
    \begin{tabular}{l|ccc}
    \toprule \toprule
        Datasets & MathVista* & MMMU* & MMMU-Pro* \\
    \midrule
\; Baseline & 83.6 & 85.8 & 77.9 \\
    \midrule
\;-- w/o $\calM_S$ (stage-1 exemplar only) & 87.2 \green{+3.6} & 85.8 \green{+0} & 80.9 \green{+3.0} \\
\;-- w/o $\calM_E$ (stage-2 schema only) & 86.4 \green{+2.8} & 86.5 \green{+0.7} & 81.3 \green{+3.4} \\
    \midrule
        Full method & 88.7 \green{+5.1} & 89.2 \green{+3.4} & 86.1 \green{+8.2} \\
    \bottomrule \bottomrule
    \end{tabular}
    \label{tab:ablation-qwen}
\end{table}

\section{Full Prompts}
\label{sec:appendix:prompts}

This section reproduces every prompt template that drives DG-Mem in verbatim form, so that the pipeline of \S\ref{sec:methods} can be reimplemented without ambiguity.
Each prompt is shown as a separate box; \emph{system} prompts are highlighted in blue, \emph{user} prompts in orange, and shared/utility templates in gray.
Placeholders rendered as \ph{name} are substituted at runtime with the corresponding field (e.g.\ \ph{question} with the test problem text, \ph{rollout\_details} with the formatted rollout block defined at the end of this section).

\subsection{Solver Prompt}
\label{sec:appendix:prompts:solver}

The frozen backbone $\MLLM$ is invoked with a single, fixed system prompt that requests a step-by-step trace and a boxed final answer. The user turn carries the multimodal problem (image plus question text); the optional retrieved memory $\calM_q$ is prepended to that user turn at test time.

\begin{promptboxsys}{Solver -- System}
\begin{footnotesize}
\begin{verbatim}
Objective:
    Solve the given problem using a step by step process.

Expected Output Structure:
Step 1:
Step 2:
...
Step n: Final Answer: \boxed{answer}

Question:
\end{verbatim}
\end{footnotesize}
\end{promptboxsys}

\subsection{Concept Summarizer ($\calK$)}
\label{sec:appendix:prompts:concept}

The concept summarizer produces the per-problem key-concept signature $\calK(q)$ used both for online categorization during training and for memory retrieval at test time. It is a single user-turn prompt; no system prompt is used, and the model is instructed not to attempt a solution.

\begin{promptboxuser}{Problem Analysis -- User}
\begin{footnotesize}
\begin{verbatim}
Objective:
    Analyze the following problem to identify its subject area and the
    key concepts, principles, formulas, or laws required for its
    solution. This analysis will be used to retrieve relevant guiding
    principles from a knowledge base.

Instructions:
    - Do not solve the problem.
    - First, identify the primary subject (e.g., Physics, Chemistry,
      Biology, Mathematics).
    - Then, list the core concepts or principles involved (e.g., Newton's
      Second Law, Conservation of Energy, Stoichiometry, Pythagorean
      theorem).
    - Keep the analysis concise and focused.

Problem:
\end{verbatim}
\ph{question}
\begin{verbatim}

Output Format:
Subject: <The primary subject>
Key Concepts: <A brief list of key concepts>
\end{verbatim}
\end{footnotesize}
\end{promptboxuser}

\subsection{Question-Level Reflector ($\Refl_R$, writes to $\calM_R$)}
\label{sec:appendix:prompts:reflector_r}

The question-level reflector condenses the labeled rollout group into IF-THEN rules that populate the transient reflection store $\calM_R$. The abstraction constraints (no problem-specific numbers, options, or entities) are what make these rules safe to consolidate into the schema memory $\calM_S$ in the post-epoch step.

\begin{promptboxsys}{Question-Level Reflector -- System}
\begin{footnotesize}
\begin{verbatim}
You are an Expert Multimodal Problem-Solving Reflector. Your task is
to extract concise, generalizable IF-THEN rules from multiple
solution attempts for a single scientific or mathematical problem.
You must analyze both visual interpretation errors (e.g., misreading
diagrams, missing chart labels) and logical reasoning errors. These
rules will be clustered by concept to serve as foundational strategic
heuristics for future AI agents.
\end{verbatim}
\end{footnotesize}
\end{promptboxsys}

\begin{promptboxuser}{Question-Level Reflector -- User}
\begin{footnotesize}
\begin{verbatim}
Given the following multimodal problem, its key concepts, and multiple
solution attempts with their verification results, extract
generalizable IF-THEN rules.

<problem>
\end{verbatim}
\ph{question}
\begin{verbatim}
</problem>

<gold_answer>
\end{verbatim}
\ph{gold\_answer}
\begin{verbatim}
</gold_answer>

<key_concepts>
\end{verbatim}
\ph{key\_concepts}
\begin{verbatim}
</key_concepts>

<rollouts>
\end{verbatim}
\ph{rollout\_details}
\begin{verbatim}
</rollouts>

Instructions:
1. Analyze the rollouts to identify the critical decision points,
   visual interpretations, or logical steps that determined success
   or failure.
2. Formulate rules as actionable IF-THEN statements.
   - The "IF" part should describe a specific problem context, visual
     pattern, or conceptual trigger (derived from the <key_concepts>).
   - The "THEN" part should provide a specific visual or logical
     intervention (e.g., "THEN explicitly verify the axis scale",
     "THEN apply the Pythagorean theorem").
3. Rule Categories (Try to extract at least one of each if applicable):
   - Positive Heuristics: "IF <context>, THEN <strategy>" -- what
     specifically worked in the correct attempts.
   - Visual/Perceptual Traps: "IF <visual context>, THEN AVOID
     <perceptual pitfall>" -- visual elements the agent
     misinterpreted in failed attempts.
   - Logical/Calculation Traps: "IF <logical context>, THEN AVOID
     <reasoning pitfall>" -- logical fallacies or formula errors
     observed in failed attempts.
4. Constraints for Generalization:
   - Rules MUST be tightly connected to the provided <key_concepts>
     so they can be accurately grouped with similar problems later.
   - Keep rules to a single, concise sentence.
5. Return 2-5 rules total. Quality and transferability over quantity.

CRITICAL CONSTRAINT -- Abstraction:
Rules MUST NOT contain problem-specific numbers, option letters
(A/B/C/D), yes/no answers, variable names (unless standard formulas),
or specific entity names. If a rule mentions any value from the
gold_answer or rollouts, rephrase it abstractly.

GOOD: "IF measuring length with a ruler, THEN align the start of the
       object with the zero mark and read the nearest whole unit at
       the endpoint."
BAD:  "IF measuring length, THEN the endpoint is at the 8 cm mark."
       (contains problem-specific value)
GOOD: "IF comparing counts of object categories, THEN systematically
       enumerate each category before comparing."
BAD:  "IF counting brown bikes, THEN there are 0 brown tandem bikes."
       (contains problem-specific entity and count)

Output Format:
Return a JSON object with a single key "rules" whose value is a list
of 2-5 IF-THEN rule strings. Each rule must start with "IF " and
contain "THEN ". If no meaningful rules can be extracted, return
{"rules": []}.
\end{verbatim}
\end{footnotesize}
\end{promptboxuser}

\subsection{Compact Reflector ($\Refl_E$, writes to $\calM_E$)}
\label{sec:appendix:prompts:reflector_e}

The compact reflector produces the exemplar entry stored in $\calM_E$: a transferable \texttt{reasoning\_strategy} and a \texttt{verification\_check}. Crucially, this prompt is the only place that \emph{is} allowed to reference the rollout text directly; the schema synthesizer below sees only the abstract IF-THEN rules in $\calM_R$, never the exemplar payload.

\begin{promptboxsys}{Compact Reflector -- System}
\begin{footnotesize}
\begin{verbatim}
You are an Expert Multimodal Problem Analyst. Your task is to extract
a generalizable reasoning strategy and verification check from
multiple solution attempts. Your output will guide a different AI
agent solving a DIFFERENT problem of the same type -- it must be
abstract enough to transfer, not specific to this instance.
\end{verbatim}
\end{footnotesize}
\end{promptboxsys}

\begin{promptboxuser}{Compact Reflector -- User}
\begin{footnotesize}
\begin{verbatim}
Given a multimodal problem, its key concepts, and multiple solution
attempts, extract a transferable reasoning strategy.

<problem>
\end{verbatim}
\ph{question}
\begin{verbatim}
</problem>

<gold_answer>
\end{verbatim}
\ph{gold\_answer}
\begin{verbatim}
</gold_answer>

<key_concepts>
\end{verbatim}
\ph{key\_concepts}
\begin{verbatim}
</key_concepts>

<rollouts>
\end{verbatim}
\ph{rollout\_details}
\begin{verbatim}
</rollouts>

Instructions:
1. Analyze where correct and incorrect approaches diverge.
2. Produce a JSON object with these fields:
   - "reasoning_strategy": A step-by-step reasoning procedure that
     would work for ANY problem of this type. Describe WHAT to do
     and HOW to verify, not what the answer is.
   - "verification_check": A specific check the solver should perform
     to catch the most common error pattern (or null if all attempts
     were correct).

CRITICAL CONSTRAINTS:
- NEVER include specific numerical values, option letters (A/B/C/D),
  yes/no answers, or entity names from this problem.
- The output must work for a DIFFERENT problem with DIFFERENT numbers
  and images.
- Frame positively: describe what TO DO, not what to avoid.

GOOD examples:
- reasoning_strategy: "Align the object's starting edge with the zero
  mark, then read the position of the ending edge to the nearest
  whole unit."
- verification_check: "Re-read the ending position independently --
  ensure it is not confused with an adjacent marking."

BAD examples (contain problem-specific values -- DO NOT do this):
- reasoning_strategy: "Count 8 individual cubes and add to 30 to get 38"
- verification_check: "The endpoint is at the 8 cm mark, not the 7 cm mark"

Output Format (return ONLY this JSON, no other text):
```json
{
  "reasoning_strategy": "<generalizable step-by-step reasoning procedure>",
  "verification_check": "<specific verification check or null>"
}
```
\end{verbatim}
\end{footnotesize}
\end{promptboxuser}

\subsection{Online Concept Categorizer ($\Cat$)}
\label{sec:appendix:prompts:cat}

The categorizer is the operational implementation of the \emph{online} taxonomy growth described in \S\ref{sec:methods:construction}. It receives the running registry of categories and decides whether to assign the new problem to an existing one or to spawn a new category, returning a structured action that the registry consumes verbatim.

\begin{promptboxsys}{Online Categorizer -- System}
\begin{footnotesize}
\begin{verbatim}
You are an Expert Problem Categorizer for multimodal scientific and
mathematical problems. Your task is to assign each problem to a
concept category, either selecting an existing category or creating
a new one when no existing category is a good fit.
\end{verbatim}
\end{footnotesize}
\end{promptboxsys}

\begin{promptboxuser}{Online Categorizer -- User}
\begin{footnotesize}
\begin{verbatim}
Given a problem's key concepts and the current set of existing
categories, decide whether to assign the problem to an existing
category or create a new one.

<problem_concepts>
\end{verbatim}
\ph{key\_concepts}
\begin{verbatim}
</problem_concepts>

<existing_categories>
\end{verbatim}
\ph{existing\_categories}
\begin{verbatim}
</existing_categories>

Instructions:
1. Compare the problem's key concepts against each existing
   category's name and id.
2. If a category is a strong conceptual match (>70% overlap in domain
   and principles), assign to it.
3. If no category is a good fit, create a new one with a descriptive
   name and representative concepts.
4. When assigning to an existing category, you may update its
   representative_concepts to better reflect the expanded set of
   problems.

Output Format (return ONLY this JSON, no other text):
```json
{
  "action": "assign" or "create",
  "category_id": "<existing category_id if assign, or a new unique slug if create>",
  "category_name": "<human-readable category name>",
  "representative_concepts": ["concept1", "concept2", ...],
  "reasoning": "<1-sentence justification>"
}
```
\end{verbatim}
\end{footnotesize}
\end{promptboxuser}

\subsection{Schema Synthesizer ($\Synth$, writes to $\calM_S$)}
\label{sec:appendix:prompts:synth}

The schema synthesizer is the post-epoch consolidation step. It reads only the IF-THEN rules in $\calM_R$ within a category (never the exemplar payload) and emits the prescriptive advice list that constitutes the category's entry in $\calM_S$.

\begin{promptboxsys}{Category-Level Reflector -- System}
\begin{footnotesize}
\begin{verbatim}
You are an Expert Multimodal Knowledge Synthesizer. Your task is to
analyze a cluster of question-level IF-THEN rules within a specific
scientific or mathematical concept category. You must synthesize
these isolated experiences into coherent, highly generalizable advice
to guide future AI agents in solving problems within this category.
\end{verbatim}
\end{footnotesize}
\end{promptboxsys}

\begin{promptboxuser}{Category-Level Reflector -- User}
\begin{footnotesize}
\begin{verbatim}
You are given a concept category, its representative concepts, and a
collection of IF-THEN rules extracted from solving multimodal
problems in this category. Synthesize them into coherent,
non-redundant, and actionable advice.

<category>
\end{verbatim}
\ph{category\_name}
\begin{verbatim}
</category>

<representative_concepts>
\end{verbatim}
\ph{representative\_concepts}
\begin{verbatim}
</representative_concepts>

<question_level_rules>
\end{verbatim}
\ph{rules\_text}
\begin{verbatim}
</question_level_rules>

Instructions:
1. Analyze and group the provided rules, eliminating redundancies and
   resolving contradictions. Elevate problem-specific rules into
   general principles applicable to the entire <category>.
2. Produce a single flat list of advice items covering:
   - Visual/perceptual strategies (interpreting images, charts,
     diagrams, spatial relationships).
   - Logical/reasoning strategies (workflows, theorems, formulas,
     step-by-step approaches).
3. Each advice item must be:
   - A single, self-contained, and actionable sentence starting with "- ".
   - **Prescriptive** -- tell the solver WHAT TO DO, not just what to
     avoid. Use "Do X when you see Y" rather than "Never assume X".
   - **Specific** -- reference concrete visual patterns, mathematical
     relationships, or domain-specific checks. Avoid abstract warnings.
   - Broadly generalizable across any problem that tests the
     <representative_concepts>.
4. Return 2-4 items total. Focus on the 2-4 most critical and
   specific insights.

IMPORTANT -- quality criteria for each rule:
- BAD (too vague): "Never assume post-event birth years from visual
  cues." -- This tells the solver to not do something but gives no
  actionable alternative.
- GOOD (prescriptive): "When estimating ages from photos to compare
  against a historical date, assess each person independently and
  count anyone who appears under 80 as likely born after 1945." --
  This gives a concrete decision procedure.
- BAD (too generic): "Always verify exact values against labeled axis
  ticks." -- This applies to any chart problem and adds no insight.
- GOOD (specific): "For scatter plots with overlapping points, trace
  each data point's y-value to the nearest labeled tick mark before
  comparing against the threshold." -- This addresses a specific
  visual challenge.

5. If no meaningful advice can be extracted, return {"advice": []}.

Output Format:
Return a JSON object with a single key "advice" whose value is a list
of 2-4 advice strings. Each string must be a single, self-contained
sentence.
\end{verbatim}
\end{footnotesize}
\end{promptboxuser}

\end{document}